\documentclass{article} 
\usepackage[final]{colm2026_conference}

\usepackage{microtype}
\usepackage{hyperref}
\usepackage{url}
\usepackage{booktabs}

\usepackage{lineno}

\definecolor{darkblue}{rgb}{0, 0, 0.5}
\hypersetup{colorlinks=true, citecolor=darkblue, linkcolor=darkblue, urlcolor=darkblue}

\usepackage{latexsym}
\usepackage[T1]{fontenc}
\usepackage[utf8]{inputenc}
\usepackage{microtype}
\usepackage{inconsolata}
\usepackage{graphicx}
\usepackage{booktabs}
\usepackage{bm}
\usepackage{amsmath}
\usepackage{amsfonts}
\usepackage{amssymb}
\usepackage{mathtools}
\usepackage{tikz}
\usetikzlibrary{calc, matrix, positioning, backgrounds, fit, decorations.pathreplacing}
\usepackage{mathabx}

\let\emptyset\varnothing

\DeclareMathOperator*{\argtopb}{arg\,top-{\mathnormal{\beta}}}

\newcommand{\alphabet}{\Sigma}
\newcommand{\seqlen}{n}
\newcommand{\conlen}{l}
\newcommand{\numcon}{k}
\newcommand{\dfa}{M}

\newcommand{\lang}{L}
\newcommand{\con}{c}
\newcommand{\beam}{B}
\newcommand{\hypo}{h}
\newcommand{\Hypo}{H}
\newcommand{\bw}{\beta}
\newcommand{\state}{q}
\newcommand{\states}{Q}
\newcommand{\accepting}{F}
\newcommand{\depth}{d}
\newcommand{\str}{s}
\newcommand{\token}{\alpha}
\newcommand{\cost}{\mathcal{C}}

\newcommand{\trans}{\delta}
\newcommand{\etrans}{\trans^*}

\title{Making Grid Beam Search Less Greedy}

\author{Sean Papay \& Roman Klinger \\
Fundamentals of Natural Language Processing \\ University of Bamberg \\ Germany \\ 
\texttt{\{sean.papay, roman.klinger\}@uni-bamberg.de} \\
}

\begin{document}

\ifcolmsubmission
\linenumbers
\fi

\maketitle

\begin{abstract}
A common formalism for constraining the output of autoregressive text generation models involves \textit{lexical constraints}, words or phrases which are required to occur in the generated text.
DFA-constrained beam search and grid beam search are two widely used paradigms for decoding from autoregressive models while enforcing lexical constraints.
As the former approach requires a number of forward passes exponential
in the number
of constraint tokens, it is often dispreferred to the latter, which requires only linearly many forward calls.
However, while grid beam search achieves an exponential speedup, it does so in a manner
which does not treat all of the constraints equally.
In this paper, we demonstrate that grid beam search is biased to
incorporate easier-to-satisfy constraints first, leaving harder constraints to the end of the sequence.
This contrasts with DFA-constrained beam search, which exhibits no such bias.
To address this shortcoming, we propose fair grid beam search, a modification to grid beam search which avoids this bias while still requiring only linearly many forward passes.
Experimentally, we confirm grid beam search's bias on two constrained generation tasks, finding significant differences in how it orders constraint tokens as compared to DFA-constrained beam search and fair grid beam search.
Furthermore, we find that fair grid beam search not only fixes grid beam search's bias, but finds higher-probability strings in the process.

\end{abstract}

\section{Introduction}
Given the tremendous success large language models (LLMs) have had in tasks across natural language processing (NLP), 
a common target of NLP research is in understanding how to best use LLMs for specific tasks.
As language models define distributions over strings of natural language, while most tasks in NLP do not involve modeling arbitrary natural language,
techniques for applying
LLMs tend to involve conditioning the LLM distribution such that the resulting conditional distribution over strings can be interpreted as a distribution over
task-specific predictions.
For classification tasks like sentiment classification \citep[e.g.,][]{zhang-etal-2024-sentiment}, this might simply involve conditioning the LLM to produce the name of a valid label,
while for structured prediction tasks like parsing \citep{drozdov2023compositional}, more sophisticated approaches may be required in order to interpret the LLM's distribution over strings as a distribution over valid structures.


The most common technique in this direction is prompting,
where LLM output is conditioned on a \textit{prompt}, which usually takes the form of natural-language instructions.
However, as prompting involves conditioning by natural-language instructions,
its success relies on the natural-language understanding (NLU) capabilities of the autoregressive model.
This is often not a problem when
a)~the model exhibits strong NLU capabilities,
b)~the instructions are easy-to understand and easy-to-follow, and
c)~some probability of failure is acceptable, but, when such assumptions do not hold, other approaches to conditioning may be taken.

One such alternative is \textit{constrained decoding} \citep{anderson-etal-2017-guided, hokamp-liu-2017-lexically, post-vilar-2018-fast, lu-etal-2021-neurologic}.
In this paradigm, a formal language of valid strings is defined, and the language model is conditioned on the event that the generated string is an element of this language.
One particularly well-explored setting for constrained decoding involves \textit{lexical constraints}: constraint languages that stipulate strings must contain a particular
substring, and intersections of such languages.
In other words, a lexical constraint stipulates that a particular word or phrase must occur in the model output, and multiple such lexical constraints can be enforced simultaneously.

Grid beam search \citep{hokamp-liu-2017-lexically} and DFA-constrained beam search \citep{anderson-etal-2017-guided} are two widely used methods for decoding from autoregressive
models while enforcing lexical constraints.
Both modifications to standard beam search \citep{Lowerre1976}, these algorithms ensure that decoding hypotheses make progress towards fulfilling all constraints by maintaining multiple beams of hypotheses.
We will compare these algorithms in detail in Section~\ref{sec:prior}; for now, it suffices to note that, for $\numcon$ lexical constraints of bounded length, grid beam search requires $O(\numcon)$ autoregressive forward passes per token,
while DFA-constrained beam search requires $O(2^\numcon)$ forward passes.
As such, grid beam search enjoys significantly more popularity.

In this paper, we draw attention to an underappreciated problem with grid beam search: It is unnecessarily \textit{greedy}, and prefers to satisfy easy lexical constraints, i.e. common constraint words, 
first, leaving harder constraints until the end.
Hence, common constraint words will occur \textit{earlier} in generated sequences than rare ones.
In addition to illustrating this problem, we also present a solution: fair grid beam search, a modification to grid beam search which elimitates this bias while largely maintaining the original's efficiency. \footnote{Fair grid beam search \textit{does} introduce a pre-computation step exponential in the number of constraints, but this does not involve forward passes through the autoregressive model, and so is independent from model size.}
This leads to higher-probability strings in the end.
In order for others to easily apply our decoding method, we release our code as an open-source python library.%
\footnote{Available at \url{https://www.uni-bamberg.de/en/nlproc/resources/fairgrid/}}

In Section~\ref{sec:prior}, we discuss existing approaches to constrained decoding, with a heavy focus on DFA-constrained beam search and fair grid beam search,
which are particularly relevant as background to our current work.
Section~\ref{sec:whygreedy} provides conceptual argumentation for the greedy bias of grid beam search.
Section~\ref{sec:fairgrid} presents fair grid beam search as an alternative to grid beam search that avoids this bias.
In Section~\ref{sec:experiments}, we present two experiments which empirically compare DFA-constrained beam search, grid beam search, and fair grid beam search, empirically
validating our claims about grid beam search's ordering bias and about fair grid beam search's improvements.
Finally, we conclude the paper in Section~\ref{sec:conclusion}.

\section{Constrained decoding}
\label{sec:prior}
We now discuss previous literature on constrained decoding in autoregressive models.
We focus on \citet{anderson-etal-2017-guided} and \citet{hokamp-liu-2017-lexically}, as we directly build on these prior works.

\subsection{DFA-constrained beam search}
We paraphrase DFA-constrained beam search as it was introduced in \citet{anderson-etal-2017-guided}.
While the approach was simply termed ``constrained beam search'' in this original publication, we refer to it as DFA-constrained beam search for the sake of disambiguity.

We begin with the observation that languages induced by sets of lexical constraints are regular.
Let $\alphabet$ be our alphabet of tokens, and let $\bm{\con} = \langle \con^1, \con^2, \cdots, \con^\conlen \rangle \in \alphabet^*$ be a lexical constraint -- a token sequence
which must occur in all generated strings.
The regular expression $\alphabet^*\con^1\con^2\cdots \con^\conlen\alphabet^*$ describes exactly those strings which satisfy the constraint $\bm{c}$.
As regular languages are closed under intersection, combinations of lexical constraints are similarly regular:
For any finite set of $k$ lexical constraints $\{\bm{\con}_1, \bm{\con}_2, \cdots, \bm{\con}_\numcon\}$, we can define a regular language
\[
L = \bigcap_i \left(\Sigma^*c_i^1c_i^2\cdots c_i^l\Sigma^*\right)
\]
where a string $\bm{s} \in \lang$ iff $\bm{s}$ satisfies all lexical constraints.

Let $\dfa = (\states, \alphabet, \trans, \state_0, \accepting)$ be a minimal DFA for $\lang$.
Without an explicit construction, it can be seen that, in the worst case, $\dfa$ requires a number of states
exponential in the number of constraints, as the automaton must ``remember'' at every time step which subset of constraints has already been satisfied.

Given such a minimal automaton $\dfa$, DFA-constrained beam search associates with each DFA state $q \in Q$ a beam of hypotheses $\beam^t(q) \subseteq \alphabet^t$, indexed by time step $t$.
Conceptually, each hypothesis is a token sequence which could potentially form a prefix to the final decoded string.
The beams are initialized to $\beam^0(\state) = \{\bm{\varepsilon}\}$ for the initial state $\state = \state_0$, and $\beam(\state) = \emptyset$ for all other $\state \neq \state_0$.

During each decoding step $t$, new hypotheses $\bm{\hypo} \in \alphabet^t$ are built by considering every possible continuation token $\token$ for each existing hypothesis $\bm{\hypo}' \in \alphabet^{t-1}$ across all beams, forming a hypothesis set
\[
\Hypo^t = \{\bm{\hypo}'\alpha \mid \bm{\hypo}' \in \bigcup_\state \beam^{t-1}(\state); \alpha \in \alphabet\}.
\]
Each such $\bm{\hypo} \in \Hypo^t$ can be assigned to a DFA state $\etrans(\state_0, \bm{\hypo}) \in Q$, where $\etrans$ is $\dfa$'s extended transition function. 
From this, we can define state-wise hypothesis sets
\[
  \Hypo^t(\state) = \{\bm{\hypo} \mid  \bm{\hypo} \in \Hypo^t, \etrans(\state_0, \bm{\hypo}) = \state\}
\]
of all hypotheses $\bm{\hypo}$ which \textit{could} be elements of the beam $\beam^t(\state)$.
To obtain the actual beam, we simply retain the top $\bw$ hypotheses in that state-wise set according to our autoregressive model's probability function $P(\bm{\hypo})$:
\[
\beam^t(\state) = \argtopb_{\bm{\hypo} \in \Hypo^t(\state)} P(\bm{\hypo})
\]
Since we would only like to generate strings in $\lang$, we only consider candidate strings originating from accepting states' beams as model output.
As with standard beam search, we proceed until our best candidate string is more probable than our highest-probability hypothesis.

\subsection{Grid beam search}
\label{sec:gbs}
Grid beam search, as it was first presented in \citet{hokamp-liu-2017-lexically}, was described in terms of an interface for building new hypotheses with
functions \texttt{generate}, \texttt{start}, and \texttt{continue}.
We reanalyze Hokamp and Liu's algorithm in terms of finite-state automata: in particular, the algorithm describes a) a method for constructing a
non-deterministic finite-state automaton (NFA), and b) a decoding procedure, analogous to DFA-constrained beam search, for generating strings accepted by this NFA.
Importantly, this decoding method, when applied to the NFA obtained from a), leads to a time complexity linear in the number of constraint tokens, as contrasted with DFA-constrained beam search's exponential runtime.

In this work, we will consider b) to be the central ``essence'' of grid beam search, and treat a) to be an implementation detail.
In fact, we note that grid beam search's decoding scheme is equally applicable to the minimal DFA for the constraint language $\lang$, with identically linear time complexity.
In light of this, we will only discuss grid beam search's decoding procedure here, and, for the remainder of this paper, we will consider grid beam search to operate on a minimal DFA for $\lang$.
In Appendix~\ref{sec:nfa}, we discuss the specific NFA constructed by \citet{hokamp-liu-2017-lexically}, and discuss how to map our construction to nondeterministic automata.

We start with the DFA $\dfa$.
We assign to each state $\state$ a \textit{depth} $\depth(\state) \in \mathbb{N}\cup\{\infty\}$ equal to the minimum number of transitions needed to reach an accepting state starting from that state (with accepting states being assigned a depth of zero, and co-inaccessible states being assigned infinite depth).
Note that this notion of depth is equivalent to the notion of constraint coverage presented in \citet{hokamp-liu-2017-lexically}, in that both measure the minimal number of tokens which must be generated before all constraints will be satisfied.

Decoding can then be defined similarly to DFA-constrained beam search, with one major difference: instead of maintaining one beam of hypotheses $\beam(\state)$ for each \textit{state},
we instead maintain one beam of hypotheses $\beam(\depth)$ for each \textit{distinct depth value}.
The beams are initialized to $\beam^0(\depth) = \{\bm{\varepsilon}\}$ for the depth of initial state $d = d(\state_0)$, and $\beam(\depth) = \emptyset$ for all depth values $\depth \neq \depth(\state_0)$.
Hypothesis sets are defined equivalently as 
\[
\Hypo^t = \{\bm{\hypo}'\alpha \mid \bm{\hypo}' \in \bigcup_\depth \beam^{t-1}(\depth); \alpha \in \alphabet\}.
\]
However, instead of defining state-wise hypothesis sets, we define depth-wise hypothesis sets
\[
  \Hypo^t(\depth) = \{\bm{\hypo} \mid  \bm{\hypo} \in \Hypo^t, \depth(\etrans(\state_0, \bm{\hypo})) = \depth\} .
\]
Forming beams proceeds identically by selecting the $\beta$ best hypotheses from each of these hypothesis sets:
\[
\beam^t(\depth) = \argtopb_{\bm{\hypo} \in \Hypo^t(\depth)} P(\bm{\hypo}).
\]
While DFA-constrained beam search required we pick our candidate strings from accepting states' beams, in grid beam search,
we select candidates from the beam for depth zero.

This construction maps the exponentially-many states of $\dfa$ to a linearly-many equivalence classes over states, and only maintains one beam for each equivalence class.
By specifying these equivalence classes in terms of depth, it is ensured that each partial candidate from each beam will have at least one child in a beam of lower depth, inductively ensuring that some candidates satisfying all constraints will be found.

\subsection{Other decoding algorithms for constraint enforcement}
In addition to the two works presented above, a number of other approaches have been proposed for modifying beam search for constrained decoding.
Two particularly prominent examples of this are \citet{post-vilar-2018-fast}, who modify grid beam search by varying beam sizes dynamically, and \citet{lu-etal-2021-neurologic}, who present a beam-search-based algorithm for decoding under unions, intersections, and negations of lexical constraints.
Apart from beam search, other approaches for constrained decoding from autoregressive models include transformations of the task into optimization in a continuous space \citep{kumar2021controlled,dathathriplug} and text generation via Metropolis-Hastings sampling \citep{miao2019cgmh}.

\subsection{Constraints from models' understanding of natural language}
With large language models, another avenue becomes available for constrained decoding: asking the model itself to enforce constraints,
relying on the model's natural language understanding capabilities.
This approach can be subdivided into prompt-based strategies and reasoning-based strategies.
When constraining via prompting, a prefix prompt specifically instructs or otherwise encourages the model to generate a text satisfying the desired constraints, which are rendered in natural langugage.
While this approach is ubiquitous in the application of LLMs to tasks with restricted output spaces,
such as classification \citep{wang-etal-2022-automatic, sanh2022multitask} and named entity recognition \citep{Ashok2023PromptNERPF, sanh2022multitask},
it is also widely applied to settings with freer responses, such as text generation with lexical constraints \citep{lin-etal-2020-commongen}.
In reasoning-based approaches, models are allowed to freely generate reasoning traces before producing an output sequence,
and may use this reasoning to strategize about how they should produce their output to best satisfy the constraints.
This approach is particularly helpful for difficult-to-satisfy constraints such as
formal theorem proving \citep{wang-etal-2024-theoremllama, pmlr-v267-wang25cb} where outputs are constrained to be valid proofs in a formal language such as Lean.
Reasoning approaches can also be combined with strict constraint enforcement, as in \cite{pmlr-v267-banerjee25a}, wherein unconstrained reasoning is followed by gramamr-constrained generation.

\section{The greedy bias of grid beam search}
\label{sec:whygreedy}
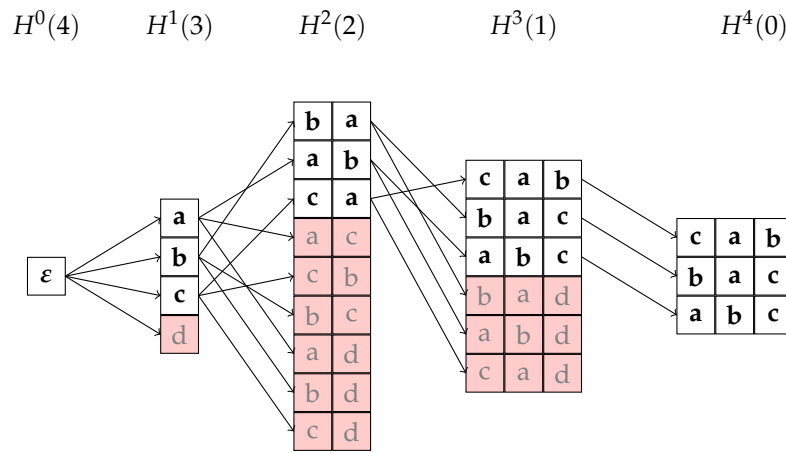
\begin{figure*}
\begin{center}
\begin{tikzpicture}
\tikzset{
    cell/.style={
        draw,
        inner ysep=0pt, 
        inner xsep=4pt,
        outer sep=0pt, 
        minimum size=.5cm, 
        anchor=center
    }
}

\tikzset{
    rowred/.style={fill=red!20, draw=none}
}

\matrix[matrix of nodes, nodes=cell, column sep=0pt, row sep=0pt] (t0) {
   $\bm{\varepsilon}$\\
};

\matrix[matrix of nodes, nodes=cell, column sep=0pt, row sep=0pt, right=of t0] (t1) {
   \textbf{a}\\
   \textbf{b}\\
   \textbf{c}\\
   \textcolor{gray}{d}\\
};
\begin{pgfonlayer}{background}
    \node[rowred, fit=(t1-4-1)(t1-4-1), inner sep=0pt] {};
\end{pgfonlayer}

\draw[->] (t0-1-1.east) -- (t1-1-1.west);
\draw[->] (t0-1-1.east) -- (t1-2-1.west);
\draw[->] (t0-1-1.east) -- (t1-3-1.west);
\draw[->] (t0-1-1.east) -- (t1-4-1.west);

\matrix[matrix of nodes, nodes=cell, column sep=0pt, row sep=0pt, right=of t1] (t2) {
   \textbf{b} & \textbf{a}\\
   \textbf{a} & \textbf{b}\\
   \textbf{c} & \textbf{a}\\
   \textcolor{gray}{a} & \textcolor{gray}{c}\\
   \textcolor{gray}{c} & \textcolor{gray}{b}\\
   \textcolor{gray}{b} & \textcolor{gray}{c}\\
   \textcolor{gray}{a} & \textcolor{gray}{d}\\
   \textcolor{gray}{b} & \textcolor{gray}{d}\\
   \textcolor{gray}{c} & \textcolor{gray}{d}\\
};
\begin{pgfonlayer}{background}
    \node[rowred, fit=(t2-4-1)(t2-9-2), inner sep=0pt] {};
\end{pgfonlayer}

\draw[->] (t1-1-1.east) -- (t2-2-1.west);
\draw[->] (t1-1-1.east) -- (t2-4-1.west);
\draw[->] (t1-1-1.east) -- (t2-7-1.west);

\draw[->] (t1-2-1.east) -- (t2-1-1.west);
\draw[->] (t1-2-1.east) -- (t2-6-1.west);
\draw[->] (t1-2-1.east) -- (t2-8-1.west);

\draw[->] (t1-3-1.east) -- (t2-3-1.west);
\draw[->] (t1-3-1.east) -- (t2-5-1.west);
\draw[->] (t1-3-1.east) -- (t2-9-1.west);

\matrix[matrix of nodes, nodes=cell, column sep=0pt, row sep=0pt, right=of t2] (t3) {
   \textbf{c} & \textbf{a} & \textbf{b}\\
   \textbf{b} & \textbf{a} & \textbf{c}\\
   \textbf{a} & \textbf{b} & \textbf{c}\\
   
   \textcolor{gray}{b} & \textcolor{gray}{a} & \textcolor{gray}{d}\\
   \textcolor{gray}{a} & \textcolor{gray}{b} & \textcolor{gray}{d}\\
   
   \textcolor{gray}{c} & \textcolor{gray}{a} & \textcolor{gray}{d}\\
};

\begin{pgfonlayer}{background}
    \node[rowred, fit=(t3-4-1)(t3-6-3), inner sep=0pt] {};
\end{pgfonlayer}

\draw[->] (t2-1-2.east) -- (t3-2-1.west);
\draw[->] (t2-1-2.east) -- (t3-4-1.west);

\draw[->] (t2-2-2.east) -- (t3-3-1.west);
\draw[->] (t2-2-2.east) -- (t3-5-1.west);

\draw[->] (t2-3-2.east) -- (t3-1-1.west);
\draw[->] (t2-3-2.east) -- (t3-6-1.west);

\matrix[matrix of nodes, nodes=cell, column sep=0pt, row sep=0pt, right=of t3] (t4) {
   \textbf{c} & \textbf{a} & \textbf{b} & \textbf{d}\\
   \textbf{b} & \textbf{a} & \textbf{c} & \textbf{d}\\
   \textbf{a} & \textbf{b} & \textbf{c} & \textbf{d}\\
};

\draw[->] (t3-1-3.east) -- (t4-1-1.west);
\draw[->] (t3-2-3.east) -- (t4-2-1.west);
\draw[->] (t3-3-3.east) -- (t4-3-1.west);

\node[above=3cm of t0.center] {$\Hypo^0(4)$};
\node[above=3cm of t1.center] {$\Hypo^1(3)$};
\node[above=3cm of t2.center] {$\Hypo^2(2)$};
\node[above=3cm of t3.center] {$\Hypo^3(1)$};
\node[above=3cm of t4.center] {$\Hypo^4(0)$};

\end{tikzpicture}
\end{center}
\caption{
\label{fig:greedy}
An illustration of the greedy bias in grid beam search with beam width 3.
Consider the following setting:
There are four constraint tokens: a, b, c, and d.
Unigram probabilities are ordered $p(\text{a}) > p(\text{b}) > p(\text{c}) > p(\text{d})$,
but strings starting with low-frequency tokens are slightly more probable than those starting with high-frequency tokens.
Assuming no non-constraint tokens, our highest-probability string should be ``dcba.''
As grid beam search discards early hypotheses with low-frequency constraint tokens, it fails to find this global optimum, and prefers hypotheses which save the hardest constraint `d' for last.
For clarity, we only illustrate the generation of constraint tokens, not non-constraint tokens.
This results in a depiction of a ``diagonal slice'' of the grid of beams, where time step and beam depth vary together.
}
\end{figure*}

We claim that grid beam search suffers a ``greedy'' bias in that it prefers hypotheses which incorporate frequent constraint tokens before infrequent ones, at the expense of average hypothesis log-likelihood.%
\footnote{This claim is not entirely novel to the present work -- \citet{lu-etal-2021-neurologic} allude to this bias ``towards sequences satisfying constraints
greedily'' in their discussion of grid beam search. However, to our knowledge, no other works have investigated this bias in depth, nor presented a direct modification to grid beam search to account for it.}
In this section we present a conceptual argument as to why this occurs.
We will make this argument in terms of competition between hypotheses which occurs during beam search, wherein the retention of a given hypothesis from one step to the next depends on the set of alternative hypotheses present in the same beam.

Suppose for simplicity that all constraints are single-token lexical constraints.
In DFA-constrained beam search, since a beam is maintained for each DFA state, there is only competition between hypotheses which have have satisfied exactly the same set of constraints.
For grid beam search, by contrast, there is competition between hypotheses which have completed the same \textit{number} of constraints, but different hypotheses may have completed different sets of constraints.

Figure~\ref{fig:greedy} illustrates how this competition can lead to decoding bias.
As words' frequencies vary considerably in natural language \citep{zipf1949human,clauset2007powerlaw}, some of our constraint words will surely be more common than others.
While specific contexts will contribute a large amount of variance, these word frequencies should affect the average autoregressive probabilities assigned to our constraint words, with common constraint words receiving higher mean autoregressive probabilities than rare ones.
On average, we would expect hypotheses including rare constraint tokens to be assigned a lower model probability than hypotheses including only common constraint tokens.

Under grid beam search, we maintain a beam for each depth, and within each beam there is competition between hypotheses which have completed the same number of constraint tokens.
Thus, in the presence of constraint tokens of varying frequency, for early (high-depth) beams, there will be direct competition between hypotheses containing only common constraint tokens, and those containing rare constraint words.
The hypotheses containing rare constraint words will tend to lose this competition, and these early beams will preferentially fill with hypotheses satisfying only easy constraints, leaving rare constraint words for later.

Another perspective on this problem is that, when comparing hypotheses in grid beam search, we only account for the likelihood of the tokens that have already been generated,
and not the likelihood of tokens that are yet to come.
While this is true of beam search in general,\footnote{In fact, this can be seen as the central simplifying assumption that separates left-to-right
approximate decoding schemes from exact decoding.} in constrained decoding, we have a priori knowledge about what tokens are yet to come, and we can take advantage of this
knowledge by rewarding hypotheses which will likely have an easier time completing the remainder of their constraints.
From this perspective, the tendency of grid beam search to complete easy constraints first is merely a symptom of an underlying inefficiency in finding high-likelihood candidates.

In Section~\ref{sec:experiments}, we will demonstrate both of these observations empirically, namely, grid beam search \textit{does} prefer to satisfy easy constraints first, and this \textit{does} lead it to finding strings with lower average log-likelihood.
But first, we will propose a simple fix to make grid beam search less greedy.

\section{Methods: fair grid beam search}
\label{sec:fairgrid}
In this section, we present a variant of grid beam search which fixes the bias discussed above.
As this bias manifests as an unfair preference for incorporating high-frequency constraints earlier, we term our variant \textit{fair grid beam search}.

\subsection{Construction}
Our construction is largely similar to that for grid beam search, with one key difference -- we associate with each state $\state$ of $\dfa$ a cost $\cost(\state)$,
representing the expected difficulty of reaching an accepting state from $\state$, and account for this cost when comparing hypotheses.
We accomplish this by weighting all arcs of $\dfa$ with $-\ln P(\token)$, the negative log unigram likelihood of that arc's symbol (token) $\token$.
We then define the cost $\cost(\state)$ of each state $\state$ to be the minimum distance to an accepting state in the underlying directed graph.
By this construction, accepting states have a cost of zero, and non-co-accessible states have infinite cost.
For notational convenience, we can also define the cost of a string $\bm{\str}$ as $\cost(\bm{\str}) = \cost(\etrans(\state_0, \bm{\str}))$.
This is simply the cost associated with the state we end at after processing $\bm{\str}$ through $M$ token-by-token, starting from the initial state.

Formally, fair grid beam search only differs from grid beam search in the construction of beams: Rather than comparing hypotheses $\bm{h}$ by probability $P(\bm{h})$,
we instead compare the quantity $\ln p(\bm{h}) - \cost(\bm{h})$:
\[
\beam^t(\depth) = \argtopb_{\bm{\hypo} \in \Hypo^t(\depth)} \ln P\left(\bm{\hypo}\right)  - \mathcal{C}\left(\bm{h}\right).
\]
As with grid-beam search, we will still be making ``unfair'' comparisons between hypotheses which have completed different subsets of constraints.
However, this cost term compensates for this by acting as a heuristic for the difficulty of completing the remainder of the constraints.
Hypotheses which have incorporated infrequent constraint tokens will have lower cost than hypotheses which have only incorporated frequent ones.

This cost-to-go heuristic $\cost(\state)$ is conceptually similar to the heuristic function used in A$^*$ search \citep{astar}, with both estimating the remaining distance from a given state to completion.
However, while A$^*$ typically employs an \textit{admissible} heuristic function -- that is, one which is guaranteed to never overestimate the true cost -- $\cost(\state)$ may arbitrarily overestimate the true cost.
This is a result of $\cost(\state)$'s definition in terms of unigram probabilities, which can vary unpredictably from the context-dependent next-token probabilities seen during autoregressive decoding.
When A$^*$ search is used with an inadmissible heuristic, the search algorithm becomes approximate, losing its guarantee to find the globally optimal path \citep{russell2021artificial}.
As beam search is already an approximate algorithm, the inadmissibility of $\cost(\state)$ carries no additional consequences for formal correctness.

Our constructions requires access to unigram probabilities from our autoregressive model's distribution $P(t)$.
Importantly, these unigram probabilities are conditioned on the prompt being used as well as the constraint setting being used,
meaning we cannot rely on precomputed corpus statistics for these unigram distributions.
Luckily, these unigram probabilities can easily be obtained in practice by simply averaging all next-token distributions seen so far during beam search.
When many texts are to be generated using a similar prompt and constraint setting, this means that the first text must be generated with standard grid beam search, but
each subsequent text will be generated with unigram probabilities obtained while generating all previous texts.
Of course, the first text may then be re-generated using the unigram probabilities obtained through the entire generation process.

\subsection{Time complexity}
\label{sec:complexity}
Assume we are interested in generating sequences of a maximal length $\seqlen$, decoding with $\numcon$ lexical constraints, each of $\conlen$ tokens, and with a beam width $\bw$.
Treating our autoregressive model as a constant-time oracle, grid beam search has a time complexity linear in all of these quantities, i.e.\ $O(\seqlen\conlen\bw\numcon)$.
This can be seen by noting that, for each of $\seqlen$ time steps, we maintain $\conlen\times\numcon$ beams, each containing $\bw$ hypotheses, and that we carry out one autoregressive call for each hypothesis that is part of a beam. 
Conversely, DFA-constrained beam search has a time complexity exponential in the number of constraint tokens, $O(\seqlen\conlen\bw2^{\numcon})$:
At each time step, we maintain one beam of $\bw$ hypotheses for each DFA state, but in the worst case we may have $\conlen\times2^{\numcon}$ states -- each state must remember which of the $2^{\numcon}$ subsets of constraints has already been satisfied,
and must remember how many of the $\conlen$ tokens of the currently-in-progress constraint have been completed.

If the state costs $\mathcal{C}$ are precomputed, fair grid beam search's time complexity is identical to that of grid beam search: the only appreciable difference is the need to obtain hypothesis
costs each time step, and this can be done in constant time if the state associated with each hypothesis is cached (constant-time determination of hypothesis state from the parent's state, and constant-time lookup of the cost from the state).
However, the precomputation of the costs $\mathcal{C}$ involves solving a single-source shortest path problem on a graph of $\left|V\right| = \conlen\times2^{\numcon}$ vertices.
Using Dijkstra's algorithm with a Fibonacci heap \citep{10.1145/28869.28874}, this can be done in $O\left(\left|V\right|\log\left|V\right|\right)$, giving the entire
algorithm a time complexity of 
\[
O\left(\left|V\right|\log\left|V\right| + \seqlen\conlen\bw\numcon\right)
= O\left(\conlen2^\numcon \left(\numcon +\log \conlen\right) + \seqlen\conlen\bw\numcon\right).
\]

This is, of course, ultimately exponential in the number of constraints, and so appears to offer no benefit over DFA-constrained beam search.
However, only the precomputation of $\mathcal{C}$ is exponential-time, and this precomputation does not involve any calls to the autoregressive model.
Thus, in typical use cases, where GPU-based model inference capacity is the primary limiting resource, this CPU-based precomputation step is unlikely to be a limiting factor to model throughput for relatively small numbers of constraints.

\section{Experiments}
In this section, we discuss two experiments we carry out which validate three claims:
\begin{description}
\setlength\itemsep{0mm}
\item[a)] grid beam search, compared to DFA-constrained beam search, tends to satisfy easy constraints first,
\item[b)] fair grid beam search corrects this bias, and
\item[c)] fair grid beam search finds higher-probability strings than grid beam search.
\end{description}
\label{sec:experiments}
Our first experiment, performed with a large number or randomly-generated constraint settings,
directly tests and validates all three claims, albeit in a somewhat artificial setting,
while our second experiment, performed at
a smaller scale with manually-curated constraint sets, compares grid beam search to fair grid beam search in a more naturalistic constrained decoding setting,
and further validates our affirmative answers to claims b) and c).
\subsection{LLM generation with many random constraints}
\label{sec:randcons}
Our first experiment is designed to highlight the tendency for grid beam search to incorporate common constraint tokens before rare ones, as
contrasted with DFA-constrained beam search, which does not exhibit this bias, and fair grid beam search, which corrects for this bias.
This experiment involves generating short texts from the TinyLlama-1.1B language model \citep{zhang2024tinyllama} with randomly chosen lexical constraints.
For each generation task, five English words are selected uniformly randomly from the 5000 most frequent words in the Corpus of Contemporary American English (COCA) \citep{COCA}.
In order to make the task setting conceptually simple and ease the interpretability of results,
we limit ourselves to constraint words which map to a single model token, reselecting whenever we choose a multi-token word.
The model is then prompted to ``write a one-sentence story,'' with no information about the constraint words provided in the prompt.
We select 1000 such five-word constraint sets, and, for each set, generate a sentence using the three constrained decoding methods.
Inference is performed in four independent ``runs'' of 250 generation tasks each -- this detail is of consequence for grid beam search, where unigram statistics are collected separately for each of these independent runs.

We are interested in the correlation between constraint word frequency and relative position within the generated sentence: we hypothesize that grid beam search should exhibit a negative correlation (constraint tokens with high frequency should have low average token index, and vice versa).
In order to avoid sensitivity to sentence length, we formalize this in terms of a notion of \textit{relative index} -- within each generated sentence, the first constraint token to appear is assigned a relative index of 1, the second a relative index of 2, and so forth, up to 5 for the final constraint token to appear.
Then, for each constrained decoding method, we can analyze the Spearman rank correlation $\rho$ between constraint token frequency and relative index across all 1000 generated texts.

\begin{table}

\begin{centering}

\begin{tabular}{lrr}
\toprule
Method & $\rho$ & $H$\\
\midrule

DFA-constrained & (0.0180) & 58.40\\
Grid & $-0.100$ & 60.81\\
Fair grid & (0.0271) & 60.48\\

\bottomrule

\end{tabular}

\begin{tikzpicture}[overlay, remember picture]

\end{tikzpicture}

\caption{\label{tab:correlations}
Results for our experiments with random constraint words.
For DFA-constrained beam search, grid beam search, and fair grid beam search, we report Spearman correlation coefficients $\rho$ between word frequency and relative index, and decoding entropy $H$.
Parenthesized correlations are found to be statistically insignificant at $p < 0.05$, while the correlation for grid beam search is significant at $p < 10^{-11}$.
A two-tailed paired $t$-test found significant pairwise differences ($p<0.0001$) between all three methods' decoding entropies.
}

\end{centering}

\end{table}

Table~\ref{tab:correlations} shows that, as hypothesized, grid beam search exhibits a small, yet significant ($p < 10^{-11}$) negative correlation: on average, low-frequency constraint words come later
than high-frequency constraint words in sequences.
For DFA-constrained beam search and fair grid beam search, we find no significant correlation at a significance level of $p < 0.05$.

In addition to investigating correlations, we also compare the model-assigned probabilities $P(\bm{s})$ of the decoded strings $\bm{s}$.
As all methods act as decoding layers for the same underlying autoregressive model, we can compare these probabilities to directly compare
how well these methods do at constrained decoding -- better decoding methods should find, on average, higher-probability strings.
We quantify this in terms of \textit{decoding entropy} ($H$), the average negative-log-probability across the full set of strings generated in a given setting. 

Decoding entropies are also listed in Table~\ref{tab:correlations}.
Across the three methods, DFA-constrained beam search achieves the lowest decoding entropy.
This is to be expected, since DFA-constrained beam search maintains a much larger set of beams, and therefore hypotheses, than the other two methods.
However, between grid beam search and fair grid beam search, fair grid beam search attains a lower decoding entropy, despite maintaining the same number of hypotheses.
A two-tailed paired $t$-test found this difference to be statistically significant ($p<0.0001$).

\subsection{CommonGen}
\label{sec:commongen}

\begin{table}[t]
\begin{center}
\begin{tabular}{lrr}
\toprule
Method & $\rho$ & $H$ \\
\midrule
Grid beam search & -0.28 & 49.22\\
Fair grid beam search & -0.06 & 48.14\\
\bottomrule
\end{tabular}
\caption{
\label{tab:commongen}
Results from our experiments on CommonGen for grid and fair grid beam search.
As before, we report Spearman correlation coefficients $\rho$ between word frequency and relative index, and decoding entropy $H$.
A two-tailed paired $t$-test found the difference between the two methods' decoding entropies to be statistically significant ($p<0.0001$).
Both correlations are significantly negative ($p<0.05$), but they differ significantly from one another as measured via bootstrapping
with a two-tailed binomial test ($p<0.00001$).
}
\end{center}
\end{table}

For our second experiment, we aim to test a more natural constrained decoding setting with a larger autoregressive model, Llama-3.1 8B Instruct \citep{grattafiori2024llama3herdmodels}.
Instead of using a random set of constraint words for each generation task, we make use of CommonGen \citep{lin-etal-2020-commongen}, a dataset designed to challenge lexically constrained generation models,
which specifies 400 distinct constrained generation tasks, each with a set of semantically-related ``concepts'' as constraints.
These concepts, formally specified as the combination of a word and a part of speech, roughly correspond to lexemes.
Thus, for example, the constraint \texttt{run\_V} could be satisfied by sentences containing the words "run," "ran," "runs," or "running."

To allow for such varied surface realizations, we use the LemmInflect library \citep{LemmInflect} to obtain a set of inflected forms for each concept, and further expand these sets by including capitalization variations.
In contrast to our previous experiment, we allow for multi-token surface realizations.
For each concept, we construct a regular expression for the the union of all surface realizations, and we take the intersection of these unions as our constraint language.
We use a minimal DFA for this constraint language to guide our decoders.

While the CommonGen task is typically framed in a setting where models are explicitly told the constraint words in a prompt, we instead choose a constraint-blind setting,
where the constraints are only enforced by the decoding scheme, and not known to the language model itself.
Although this setting precludes numerical comparisons to prior work on CommonGen, and in fact makes the task significantly harder,
it allows us to better analyze the effects of the decoding method in isolation, without any interference by the NLU capabilities of the language model.%
\footnote{
Preliminary experiments showed that including constraint concepts in the prompt significantly affected the unigram probabilities of the tokens comprising those constraints,
leading to erratic performance when using fair grid beam search with precomputed unigram probabilities.}

As with our first experiment, we compare two values across decoding schemes: the Spearman rank correlation $\rho$ between constraint word frequency and relative index, and decoding entropy $H$.
To account for varied surface realizations of constraint words, we take as word frequencies the sum of all surface realizations frequencies in COCA.
As the automata we obtain with this approach have significantly more states than those for our previous experiment,
we do not test DFA-constrained beam search in this setting, and only compare grid beam search to fair grid beam search.

Table~\ref{tab:commongen} lists our results from this experiment.
In summary, we reconfirm the two hypotheses we sought to validate with this experiment: that grid beam search preferentially satisfies easy constraints first, and that fair grid beam search
achieves a lower decoding entropy than fair grid beam search.
Of note, both grid beam search and fair grid beam search achieve a significant negative correlation coefficient, just of different magnitudes.
However, the significant ($p<0.00001)$ difference between the two correlation coefficients indicates that grid beam search introduces an additional bias not present with fair grid beam search.

The decoding methods' effect on both constraint ordering and decoding entropy appear to be much stronger in this experiment than they had been for our setting with a smaller language model and random constraints.
Grid beam search has a correlation coefficient of $-0.28$ (compared to $-0.10$ for the previous experiment), and fair grid beam search improves upon grid beam search by over one nat in decoding entropy.
This provides strong evidence that the problem we point out in grid beam search, and the solution we present with fair grid beam search, are of practical relevance
in realistic constrained decoding settings.

\subsection{Wall-clock efficiency}
As we noted in Section~\ref{sec:complexity}, fair grid beam search introduces an exponential-time precomputation step that is not needed for standard grid beam search.
Since this precomputation takes place on the CPU, and since it does not scale with model size, the usual limiting factor for LLM inference workflows, the practical consequences of this precomputation are not clear a priori.
Therefore, in this section, we briefly compare the wall-clock time usage of grid and fair grid beam search for our two experiments.

\begin{table}
\begin{center}
\begin{tabular}{lr@{ }lr@{ }l}
\toprule
& \multicolumn{2}{c}{\shortstack{Experiment 1:\\Random Constraints}} & \multicolumn{2}{c}{\shortstack{Experiment 2:\\CommonGen}}\\
\midrule
\textbf{Grid beam search} & 21.23 s & & 214.65 s\\
\textbf{Fair grid beam search} & 21.99 s & & 208.48 s\\
\hspace{1em}Precomputation time & 0.57 s & (2.6\%) & 0.93 s & (0.9\%)\\
\hspace{1em}Decoding time & 21.42 s & (97.4\%) & 206.5 s & (99.1\%)\\
\textbf{DFA-constrained beam search} & 46.59 s & & ---\\
\bottomrule

\end{tabular}
\end{center}
\caption{\label{tab:wall} Average wall-clock times to generate a single text across our experiments and decoding methods, in seconds.
For fair grid beam search, we decompose this in terms of time spent on the precomputation step and time spent on beam-search decoding.
}
\end{table}

Table~\ref{tab:wall} lists the average wall-clock time per generated string across our two experiments and decoding methods.
We find that, for these experimental settings, the precomputation step takes less than three percent of the total time requirements for text-generation, and that the average wall clock time requirements for grid beam search and fair grid beam search are largely comparable.
Of course, with a sufficiently large number of constraints, the time for the precomputation step \textit{would} come to dominate the decoding time, but this does not occur in the constraint settings investigated in these experiments.

\section{Conclusion}
\label{sec:conclusion}
In this work, we discuss the greedy bias of grid beam search and how to fix it.
We present a conceptual argument as to why grid beam search preferentially satisfies easy constraints first, and experimentally demonstrate that it in fact does so in practice, biasing the constraint orderings of generated sentences.
To correct for this bias, we present fair grid beams search, a slight modification to the decoding algorithm that rewards hypotheses for including difficult constraints.
Experimentally, we demonstrate that fair grid beam search not only fixes grid beam search's bias in constraint ordering, but that it finds higher-probability strings in the process.
While the time complexity of fair grid beam search is worse than grid beam search, all extra computation is in a precomputation step that requires no access to the underlying autoregressive model,
meaning that the improvements provided by fair grid beam search essentially come ``for free'' in settings where model throughput is the computational bottleneck.


\section*{Acknowledgements}
This work is funded by the project INPROMPT
(Interactive Prompt Optimization with the Human
in the Loop for Natural Language Understanding
Model Development and Intervention, funded by
the German Research Foundation, KL 2869/13–1,
project no. 521755488).

The authors gratefully acknowledge the scientific support and HPC resources provided by the
Erlangen National High Performance Computing Center (NHR@FAU) of the Friedrich-Alexander-Universität Erlangen-Nürnberg (FAU)
under the BayernKI project v121ca.
BayernKI funding is provided by Bavarian state authorities.

\bibliography{lit}
\bibliographystyle{colm2026_conference}

\appendix
\label{sec:nfa}

\section{Grid beam search as a finite-state automaton}
\citet{hokamp-liu-2017-lexically} define grid beam search in terms of an interface of three functions: \texttt{generate}, \texttt{start}, and \texttt{continue},
and in terms of hypotheses, which may either be \textit{open} (not ``working on'' any constraint) or \textit{closed} (currently ``working on'' one particular constraint.
We reinterpret this description as describing the construction of a nondeterministic finite-state automaton.
We take the states of our automaton to be complete answers to the following set of questions:
\begin{description}
  \item[a)] Which subset of constraints has already been completed?
  \item[b)] Which constraint, if any, is currently being worked on?
  \item[c)] If applicable, how many tokens of it have already been completed?
\end{description}
That is, each state $q_i$ is a triple of answers to questions a), b), and c).
For a constraint set $C$, the starting state, $q_1$, answers these questions as ($\emptyset$, none, not applicable), and the sole accepting state answers these questions ($C$, none, not applicable).

The three functions \texttt{generate}, \texttt{start}, and \texttt{continue} define the transitions of the automaton.
For any possible subset of constraints $A \in 2^C$, and for any symbol $\alpha \in \Sigma$, \texttt{generate} defines the self loop transition
\[
(A, \text{none}, \text{not applicable}) \xrightarrow{\alpha} (A, \text{none}, \text{not applicable}).
\]

For a constraint $c_i \notin A$, \texttt{start} defines transitions of the form
\[
(A, \text{none}, \text{not applicable}) \xrightarrow{c_i^1} (A\cup\{c_i\}, \text{none}, \text{not applicable})
\]
when $c_i$ is a single token constraint, and 
\[
(A, \text{none}, \text{not applicable}) \xrightarrow{c_i^1} (A, c_i, 1)
\]
otherwise.

Finally, \texttt{continue} generates transitions of the form 
\[
(A, c_i, j-1) \xrightarrow{c_i^{j}} (A\cup\{c_i\}, \text{none}, \text{not applicable})
\]
when $\left|c_i\right| = j$, and 
\[
(A, c_i, j-1) \xrightarrow{c_i^{j}} (A, c_i, j)
\] otherwise.

This automaton is non-deterministic: for each state not currently working on a constraint, \texttt{generate} defines one outgoing transition
for every token in our vocabulary, while \texttt{start} defines distinct transitions for some tokens.
Thus, the construction we present in Section~\ref{sec:gbs} cannot be directly applied to this automaton.
This is not a fundamental difficulty, but rather a notational mismatch.
In fact, the only change that needs to be made is to modify our definition of the depth-wise hypothesis sets to
\[
  \Hypo^t(\depth) = \{\bm{\hypo} \mid  \bm{\hypo} \in \Hypo^t, \exists \state_i: \state_i \in \etrans(\state_0, \bm{\hypo}) \wedge \depth(\state_i) = \depth\}
\]
in order to account for the automaton's transition function $\trans$, and consequently $\etrans$, being set-valued instead of state-valued.

Conceptually, this does change the picture, in that hypotheses can now exist in multiple beams, rather than just one.
Concretely, when generating the first token of a constraint, this can either be done via the \texttt{start} transition (in which case that token is ``counted'' as the start of a constraint,
or via the \texttt{generate} transition (in which case it isn't counted), leading to the same token sequence appearing as hypotheses for two distinct depth values.
While retaining two copies of the same hypothesis might leave less room in a beam for other hypotheses,
we do not expect this to be very common or consequential in practice.

\end{document}